\documentclass[11pt,a4paper]{article}
\usepackage{times,latexsym}
\usepackage{url}
\usepackage[T1]{fontenc}
\usepackage{graphicx,epstopdf}
\usepackage{multirow}
\usepackage{algorithm}
\usepackage{algorithmic}
\usepackage{amsmath}
\usepackage{amsfonts}
\usepackage{subfig}

\usepackage{tikz-dependency}
\usepackage[acceptedWithA]{tacl2018v2}

\usepackage{xspace,mfirstuc,tabulary}

\newif\iftaclinstructions
\taclinstructionsfalse % AUTHORS: do NOT set this to true
\iftaclinstructions
\renewcommand{\confidential}{}
\renewcommand{\anonsubtext}{(No author info supplied here, for consistency with
TACL-submission anonymization requirements)}
\newcommand{\instr}
\fi

\iftaclpubformat % this "if" is set by the choice of options

\else

\fi

\title{How Does Adversarial Fine-Tuning Benefit BERT?}

\author{
 Javid Ebrahimi, Hao Yang, Wei Zhang  \\
 Visa Research, Palo Alto, USA \\
  {\sf\{jebrahim, haoyang, wzhan\}@visa.com} \\
}

\date{}

\begin{document}
\maketitle
\begin{abstract}
 Adversarial training (AT) is one of the most reliable methods for defending against adversarial attacks in machine learning. Variants of this method have been used as regularization mechanisms to achieve SOTA results on NLP benchmarks, and they have been found to be useful for transfer learning and continual learning. We search for the reasons for the effectiveness of AT by contrasting vanilla and adversarially fine-tuned BERT models. We identify partial preservation of BERT's syntactic abilities during fine-tuning as the key to the success of AT. We observe that adversarially fine-tuned models remain more faithful to BERT's language modeling behavior and are more sensitive to the word order. As concrete examples of syntactic abilities, an adversarially fine-tuned model could have an advantage of up to 38\% on anaphora agreement and up to 11\% on dependency parsing. Our analysis demonstrates that vanilla fine-tuning oversimplifies the sentence representation by focusing heavily on a small subset of words. AT, however, moderates the effect of these influential words and encourages representational diversity. This allows for a more hierarchical representation of a sentence and leads to the mitigation of BERT's loss of syntactic abilities.
\end{abstract}

\section{Introduction}
Pretrained language models (LMs) \cite{devlin2018bert,peters2018deep,brown2020language,yang2019xlnet} have transformed the natural language processing (NLP) landscape, enabling researchers and practitioners alike to build far better models more readily and flexibly than in the past. Given their exceptional success, they have been the focus of expansive exploratory research in the last couple of years. These works range from more machine-learning-oriented analysis \cite{hao2019visualizing,aghajanyan2020intrinsic,zhang2021inductive} to more linguistics-driven analysis \cite{tenney2019bert,hewitt2019designing,jawahar2019does}. 
All of these efforts deal with having a better understating of what and how deep neural networks learn.

This work focuses on the analysis of adversarial training (AT) \cite{goodfellow2014explaining,madry2017towards,wong2020fast} for fine-tuning of BERT \cite{devlin2018bert}, bridging the gap between the conventional wisdom regarding AT's effectiveness on the one hand, and its implication for the linguistic abilities of BERT on the other. Based on the results presented in this paper, we conjecture that (partial) preservation of BERT's syntactic abilities is the key behind the success of AT. Our findings corroborate a hypothesis postulated by \citet{mccoy2019right}: The failure of fine-tuned BERT models on tasks, which require syntactic generalization, is due to characteristics of the downstream process, rather than BERT's inherent weaknesses.

Variants of AT have been used to achieve state-of-the-art (SOTA) results on NLP benchmarks \cite{jiang2019smart,zhu2019freelb,chen2020seqvat,wu2017adversarial,pereira2020adversarial,pereira2021targeted,yasunaga2017robust}. The inner workings of AT and how it helps training are not quite understood, though recent work suggests adversarial perturbations facilitate the learning of more general-purpose representations \cite{salman2020adversarially,aghajanyan2020better,jiang2019smart}. Specifically, AT is found to be useful for transfer learning in computer vision \cite{salman2020adversarially}, wherein this phenomenon is attributed to AT's extraction of \textit{robust features} \cite{ilyas2019adversarial}. Similarly, the effectiveness of AT for transfer learning has also been reported in NLP \cite{jiang2019smart}. \citet{aghajanyan2020better} study the accuracy of fine-tuned BERT models under a limited continual learning setting, and they find that adversarial training increases resistance to \textit{catastrophic forgetting} \cite{mccloskey1989catastrophic}. As syntax of a given language is shared for different downstream tasks, preserving the syntactic knowledge during fine-tuning could explain these empirical findings on the benefits of AT. As another piece of evidence, \citet{cheng2020posterior} demonstrate the effectiveness of AT for challenging natural language inference examples, for which syntactic generalization is arguably required. Our work puts forth a framework to explain these empirical findings, using well-known probes and novel intrinsic methods of analysis.

Parameter-free probes \cite{linzen2016assessing,marvin2018targeted} and parametric probes \cite{belinkov2017neural,hewitt2019structural} are often employed to measure linguistic abilities of a single model. We pursue a \textit{contrastive} analysis of two fine-tuning procedures using probes. We contrast vanilla (VAN) and adversarially fine-tuned (ADV) BERT models on their syntactic abilities. This includes {probing} models for linguistic phenomena such as subject-verb agreement and NLP tasks such as dependency parsing, on all of which BERT itself performs well.
While fine-tuned models have to specialize to do a specific classification task and some of BERT's capabilities are expected to be lost, we find that AT mitigates overspecialization and modulates this loss.

Probes are used to map neural representations to linguistic formalisms. However, it has been demonstrated that grammars learned by neural models differ from expert-designed ones \cite{williams2018latent,wu2020perturbed}. In addition, due to their inductive biases, deep neural networks themselves could be regarded as linguistic theories \cite{baroni2021proper}. Thus, to explain syntactic abilities of ADV models, we propose new methods to analyze models' representations intrinsically, which can reliably differentiate between ADV and VAN models, without relying on linguistic formalisms. Through our experiments on text classification, we find that AT helps BERT preserve some of its syntactic abilities by moderating the impact of individual words, preventing them from dominating the sentence representation. For instance, in a sentiment classification task, sentiment-related words tend to influence the representations much more significantly in a VAN model. Precisely,
\begin{enumerate}
 
\item Using a low-rank approximation of representations at each layer, we find that ADV models' representations are richer and harder to compress. Specifically, following a singular value decomposition, there exists a more severe degradation in the performance of ADV models, compared with VAN counterparts.   
\item
Using a gradient-based measure, which measures the influence of words over each other, we create pairwise relation graphs and extract dependency trees from them. We find that AT helps create deeper trees with lower branching factors. This explains the superiority of ADV models on syntactic tasks which are hierarchy-sensitive. We also analyze spectral properties of relation graphs and arrive at similar conclusions about their geometry.

\end{enumerate}

The rest of the paper is organized as follows: We present a background on adversarial training section in \S \ref{sec:background}. In section \S \ref{sec:proxy}, we provide the initial evidence regarding ADV models' superiority over VAN models in syntactic tasks. This is followed by sections \S \ref{sec:free} and \S \ref{sec:param}, which provide concrete results based on probes. In section \S \ref{sec:intrinsic}, we provide our intrinsic analyses which help explain the results in previous sections. 
Section \S \ref{sec:related} gives an additional overview of relevant work, and finally, we will conclude with suggestions for future work in section \S \ref{sec:conclusion}.

\section{Background}
\label{sec:background}
 Adversarial training involves creating gradient-based adversarial examples during training to defend against test-time adversaries. \citet{goodfellow2014explaining} proposed the fast gradient sign method (FGSM) to generate adversarial examples with a single gradient step.  
 Basic iterative method (BIM) \citep{kurakin2016adversarial} improved upon FGSM by iteratively taking smaller FGSM steps, leading to the finding of harder-to-defend adversarial examples. Earlier implementations of adversarial training included augmenting the dataset with the adversarial examples. \citet{madry2017towards} expanded upon BIM by using additional random restarts for the multi-step search and was the first to perform \textit{adversarial risk minimization}, excluding clean examples from training. This method is widely adopted as a strong defense and is known as the projected gradient method (PGD) in the literature.
 \subsection{PGD}
 Given a model $f_\theta$, parameters $\theta$, a dataset $\mathcal{D}$, a loss function $\ell$ and a threat model $\Delta$, the following is called the robust optimization problem, 

\begin{equation}
\label{eq:adv}
    \min_\theta \sum_{x_i, y_i \in \mathcal{D}} \max_{\delta \in \Delta} \ell(f_\theta(x_i + \delta), y_i).
\end{equation}

The threat model is typically a norm-bounded noise, e.g., $\Delta = \{\delta : \|\delta\|_2 \leq \epsilon\}$ for some $\epsilon >0$. The $\ell_2$ threat model has been used by most prior work in NLP \cite{miyato2016adversarial,jiang2019smart,zhu2019freelb} which we will use throughout the paper. 

Adversarial risk minimization for neural networks comprises approximating the inner maximization over $\Delta$, followed by gradient descent on the model parameters $\theta$. To solve the inner maximization, one can use a one-shot linear approximation using the gradients of the model at the given input. For an $\ell_2$ perturbation, the optimal noise vector has the following closed form: 
\begin{equation}
    \delta^\star = \epsilon\cdot \frac{\nabla_{x_i} \ell(f(x_i),y_i)}{\lVert \nabla_{x_i} \ell(f(x_i),y_i) \rVert}. 
\end{equation}
We could improve this approximation by taking multiple steps of size $\alpha < \epsilon$. It is likely that $\delta$ leaves the threat model, in which case it is projected back to the set $\Delta$. That is, it will either stay unchanged if $\lVert \delta \rVert \leq \epsilon$, or be projected onto the $\ell_2$ ball (i.e., $\epsilon\frac{\delta}{\lVert \delta \rVert}$) if $\lVert \delta \rVert > \epsilon$. Larger number of steps to solve the inner maximization leads to more difficult adversarial examples, and hence the improvement of the test-time robustness. 

NLP models work with discrete symbols, for which norm-bounded perturbations are not valid adversarial examples. Thus, adversarial training over the input embeddings is often used as a regularization mechanism to improve generalization rather than robustness to adversarial examples. Therefore, majority of adversarial training works in NLP use single-step approximations \cite{jiang2019smart,liu2020adversarial, miyato2016adversarial} or use only a few steps \cite{zhu2019freelb}. Furthermore,
\citet{wong2020fast} demonstrate that FGSM with random initialization of $\delta$ is as effective as PGD in defending against adversarial examples. We compared a 20-step PGD with a single-step one, both with randomization, and found the former performed better in most of our experiments. Some of these results are reported in section \S \ref{sec:proxy}. For the initialization of $\delta$, we use the scheme proposed by \citet{zhu2019freelb}\footnote{This initialization of $\delta$ is not guaranteed to remain in $\Delta$, but we empirically found it to be a good choice.}. 
See Algorithm \ref{alg:pgd}. 
\begin{algorithm}[t]
\caption{Adversarial training with an $\ell_2$ threat model, for $E$ epochs, a text dataset of size $M$ with their lengths $L$, $N$ PGD steps, a model $f_\theta$, radius $\epsilon$, and the iteration step size $\alpha$. Adapted from \cite{wong2020fast}.}
\label{alg:pgd}
\begin{algorithmic}
\FOR {$e=1\dots E$}
\FOR {$i=1\dots M$}
\STATE $\delta = \frac{1}{\sqrt{L_i}}\text{Uniform}(-\epsilon, \epsilon)$  
\FOR {$j=1\dots N$}
\STATE $\delta = \delta + \alpha \cdot \frac{\nabla_\delta \ell(f_\theta(x_i + \delta),y_i)}{\lVert \nabla_\delta \ell(f_\theta(x_i + \delta),y_i) \rVert}$
\STATE $\delta = \frac{\delta}{\text{max}(1, \frac{\lVert \delta \rVert}{\epsilon})}$
\ENDFOR
\STATE $\theta = \theta - \nabla_\theta \ell(f_\theta(x_i + \delta), y_i)$ 
\ENDFOR
\ENDFOR
\end{algorithmic}
\end{algorithm}

\section{Contrastive Analysis of Fine-Tuning}
\label{sec:probe}
We use BERT's \texttt{base-cased} model from the HuggingFace repository \cite{wolf2019huggingface}. for our analyses. We use the DBpedia ontology dataset \cite{zhang2015character}, the subjectivity analysis (SUBJ) dataset \cite{pang2004sentimental},  the AG's News dataset, and the movie review (MR) dataset \cite{pang2005seeing}. The methodology involves fine-tuning models based on vanilla and adversarial training on these datasets, and contrasting them with the base BERT model. These models are either used directly in parameter-free probes (\S \ref{sec:free}) or used to train linear or MLP-based probes (\S \ref{sec:param}).

We train our models with Adam \citep{kingma2014adam} using the initial learning rate of 2e-5 for 6000 steps with linear decay. We use 20 steps of PGD, set $\alpha$ to be 20\% of $\epsilon$, and tune the $\epsilon \in \{\text{1e-1, 2e-1, 3e-1, 4e-1}\}$. This single hyperparameter, $\epsilon$, is tuned using the linguistic probing based on \cite{marvin2018targeted}, the results of which we will report in section \S \ref{sec:proxy}. We use batch size of 32 and limit the maximum length to 128. We fine-tune our models once and, for each point of comparison, we report the average performance of ten best checkpoints (i.e., the best accuracies on the dev set).

We focus on text classification tasks where each sentence is preceded by a \texttt{CLS} symbol and succeeded by an \texttt{SEP} symbol. We avoid downstream tasks in which examples are composed of pairs of sentences. This choice was motivated by limiting our experiments to a more controlled setup, in particular avoiding the side-effects of the cross-attention mechanism, the dynamics of which might be more difficult to interpret.  

\subsection{Analysis Through Proxies} \label{sec:proxy}
To start our investigation on implications of AT, we start by two simple analyses, which we call proxy methods of analysis that motivate the rest of the experiments in the paper. The following results inform us about the possibility of preserving syntactic abilities of BERT through AT. Experiments in this section include AT with one step (ADV-1) (i.e., $\alpha=\epsilon$) and 20 steps (ADV-20).

\begin{figure*}[h]
    \centering
    \scalebox{1.0}{
    \includegraphics[width=\textwidth]{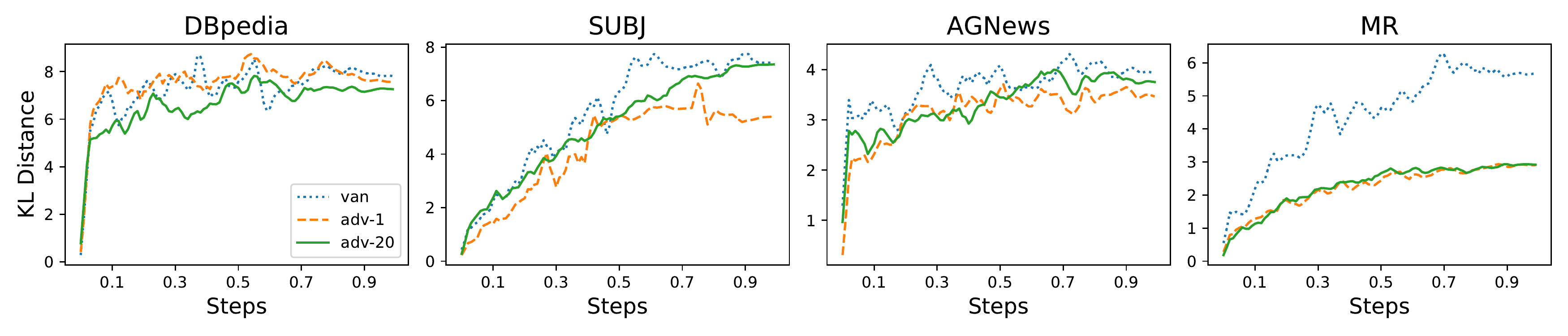}
    }
    \caption{The Symmetrized KL distance between the base BERT model and the models under fine-tuning, on a masked language modeling task using the dev set of the datasets. As training progresses, the KL distance increases, but the ADV models remain more faithful to the base model.}
    \label{fig:loss}
\end{figure*}
Pretraining objective functions directly influence linguistic knowledge of LMs \cite{zhang2021inductive,nikoulina2021rediscovery}. Naturally, we expect that fine-tuning for a specific task, trained with a different objective function, would lead to the loss of some of the capabilities they have acquired during pretraining. We measure the deviation of a fine-tuned model from the pretrained LM by measuring their symmetrized Kullback–Leibler (KL) distance (i.e., the summation of the KL divergences in both directions) in a masked language modeling (MLM) task. We simply use the dev set of each datasets for this experiment. For each example, we \textit{mask} one word at a time, compute the KL distance between the two models' predictions for the masked word, and take the average for all words of all examples in the dev set. We see in Figure \ref{fig:loss} that as training progresses, a noticeable contrast between a VAN and an ADV model emerges. With the exception of ADV-1 on DBpedia, all ADV models deviate less than the VAN model from the base model. This is particularly interesting as AT does not directly encourage proximity to the base model. 

Recent works \cite{sinha2021masked,pham2020out,gupta2021bert} have pointed to the insensitivity of BERT to the order of words, pointing at the need for better benchmarks as well as casting doubts on BERT's syntactic abilities. Table \ref{tab:order} shows the accuracy of the models on the ordered original dev set, alongside the average accuracy on ten sets with randomly-ordered words. The performance of the VAN and ADV models are similar on the original dev set, and the performance of all models degrade on randomly-ordered sets. On all of the datasets, the performance of the ADV-20 model is the lowest, with the drops being most significant on the SUBJ dataset. Specifically, the drop in this dataset is 16\% for the VAN and 25\% for the ADV-20 model. 
\begin{table}[h]
    \centering
\scalebox{0.75}{
\begin{tabular}{|c|c|c|c||c|c|c|}
\hline
& \multicolumn{3}{c|}{ordered $\uparrow$} & \multicolumn{3}{c|}{random $\downarrow$}  \\
\cline{2-7}
Dataset & van  & adv-1 & adv-20 & van  & adv-1 & adv-20       \\
 \hline
DBpedia & {99.37} & 99.30 & 99.33 & 94.57 & 94.43 & \underline{93.95}  \\
SUBJ & {96.76} & \underline{97.10} & 96.47 & 81.21 & 77.70 & \underline{72.62}  \\
AGNews & \underline{94.34} & 94.05 & 93.72 & 90.24 & 89.43 & \underline{88.63}  \\
MR & {87.05} & 86.71 & 86.94 & 71.25 & 71.90 & \underline{68.55}  \\
\hline
\end{tabular}
}
    \caption{Accuracy on the original (ordered) dev set and the average accuracy of ten corresponding sets with randomly ordered words. ADV models are more sensitive to the order of words.}
\label{tab:order}
\end{table}

Both of these proxy methods of analysis indicate that adversarial training could help BERT improve the preservation of its syntactic abilities during fine-tuning. That is, closer resemblance to the BERT base model, and less resemblance to a bag-of-words model, suggests a more syntactically-aware model.  \citet{sinha2020unnatural} also use word order sensitivity as a prerequisite for syntactic knowledge. Our work, though, emphasizes the downstream fine-tuning process as the reason for the weakness in syntactic generalization. Next, we perform comprehensive probe-based comparisons based on models' performances on syntactic tasks. 
\begin{table*}[h]
    \centering
    \scalebox{0.85}{
\begin{tabular}{|c|c|c|c|c|c|c|c|c|c|c|}
\hline
&   & \multicolumn{2}{c|}{DBpedia} &  
\multicolumn{2}{c|}{SUBJ} & \multicolumn{2}{c|}{AGNews} & \multicolumn{2}{c|}{MR}   
&   \\
\cline{2-11}
 Task & base  & van    & adv & van  & adv &  van  & adv  &  van  & adv  & {\#pairs}    \\
 \hline
Subject-Verb Agreement & 90.98 & 88.70 & \underline{89.40} & \underline{90.70} & 90.14 & 90.55 & \textbf{\underline{91.66}} & 78.22 & \underline{88.27} & 121,220  \\
Anaphora Agreement & \textbf{95.26} & \underline{95.06} & 93.54 & 73.66 & \underline{89.38} & 65.00 & \underline{90.68} & 69.48 & \underline{89.97}  & 26,040 \\
 \hline
 \hline
 Subject-Verb Agreement  & 90.88 & 90.38 & \underline{91.71} & 88.93 & \textbf{\underline{91.75}} & 90.16 & \underline{91.28} & 87.68 & \underline{90.85}   & 3,739 \\
 
Determiner Noun Agreement & \textbf{98.35} & \underline{98.16} & 97.46 & 96.01 & \underline{97.01} & 96.84 & \underline{97.91} & 92.81 & \underline{97.55}    & 6,926 \\

Anaphora Agreement  & 97.15 & 98.22 & \textbf{\underline{98.31}} & 88.90 & \underline{96.10} & 95.02 & \underline{98.08} & 89.90 & \underline{95.50} & 2,000 \\

Argument Structure & \textbf{81.61} & \underline{77.60} & 77.57 & 77.99 & \underline{79.00} & 77.89 & \underline{79.09} & 74.39 & \underline{75.74} & 3,056  \\

Irregular Forms & 97.20 & 95.00 & \underline{95.06} & 93.76 & \underline{95.88} & 95.51 & \textbf{\underline{97.33}} & 85.23 & \underline{95.15} & 1,963 \\

Island Effects &    76.40 & \underline{77.36} & 75.07 & \underline{84.40} & 84.01 & 78.79 & \textbf{\underline{85.59}} & 75.96 & \underline{81.63} & 1,000 \\

\hline
\end{tabular}
}
    \caption{Performance on syntactic and morphological probing tasks. In the overwhelming majority of the cases, ADV models perform better than their VAN models. Top two row tasks are from \cite{marvin2018targeted}, and the rest are from \cite{warstadt2020blimp}. For the latter, {we only consider tasks wherein the base model is at least 70\% accurate and at least 1,000 pairs exist.} Bold numbers denote the best in each task, and underscored numbers denote the best in each pair of comparison between VAN and ADV.}
    \label{tab:param-free-probe}
\end{table*}

\subsection{Parameter-Free Probes}
\label{sec:free}
In our first probe-based analysis, we contrast ADV\footnote{For the remainder of the paper, we use the ADV-20 model alone and simply refer to it as ADV.} and VAN models on their linguistic knowledge across a few syntactic and few morphological phenomena of English. We use two synthetically generated datasets \citep{marvin2018targeted,warstadt2020blimp}. These datasets are composed of pairs of similar English sentences, wherein one sentence of the pair has a grammatical error. We only consider pairs which are different in only one word and follow the methodology of \citet{goldberg2019assessing}, wherein we feed BERT the complete sentence and \textit{mask} the single focus
word. Then, we probe BERT by comparing the score
assigned to the correct word with the score
assigned to the incorrect one. For instance, a minimal pair for the Argument Structure from \cite{warstadt2020blimp} contains ``A teacher wasn't insulted by Julie.'' as the correct, and ``A teacher wasn't died by Julie.'' as the incorrect sentence. We feed ``A teacher wasn't \texttt{MASK} by Julie.'' and compare the score of ``insulted'' with that of ``died'' given by the language modeling layer of BERT. Given this type of evaluation, we only consider pairs with focus words which appear in the subword-based vocabulary of BERT.
 
\begin{table*}[h]
    \centering
    \scalebox{0.9}{
\begin{tabular}{|c|c|c|c|c|c|c|c|c|c|}
\hline
\multirow{2}{*}{Task} &   & \multicolumn{2}{c|}{DBpedia} &  \multicolumn{2}{c|}{SUBJ} & \multicolumn{2}{c|}{AGNews} & \multicolumn{2}{c|}{MR}  \\
\cline{2-10}
 & base  & van  & adv & van  & adv & van  & adv &  van  & adv     \\
 \hline
POSL & \textbf{94.70} & 93.62 & \underline{93.90} & 92.67 & \underline{93.33} & 91.74 & \underline{93.07} & 90.86 & \underline{93.31}  \\
DAL & \textbf{92.07} & 90.54 & \underline{91.17} & 89.72 & \underline{90.57} & 88.22 & \underline{89.78} & 87.53 & \underline{90.32}  \\
Parsing & \textbf{79.29} & 74.85 & \underline{76.39} & 72.73 & \underline{74.78} & 69.74 & \underline{73.97} & 66.59 & \underline{74.19}  \\

\hline
\end{tabular}
}
    \caption{Performance on POSL and DAL, and dependency parsing using linear probes. In all cases, an ADV model does better than its VAN counterpart. Bold numbers denote the best in each task, and underscored numbers denote the best in each pair of comparison between VAN and ADV models.}
    \label{tab:parameter-based-probes}
\end{table*}
\begin{figure*}[h]
    \centering
    \scalebox{1.0}{
    \includegraphics[width=\textwidth]{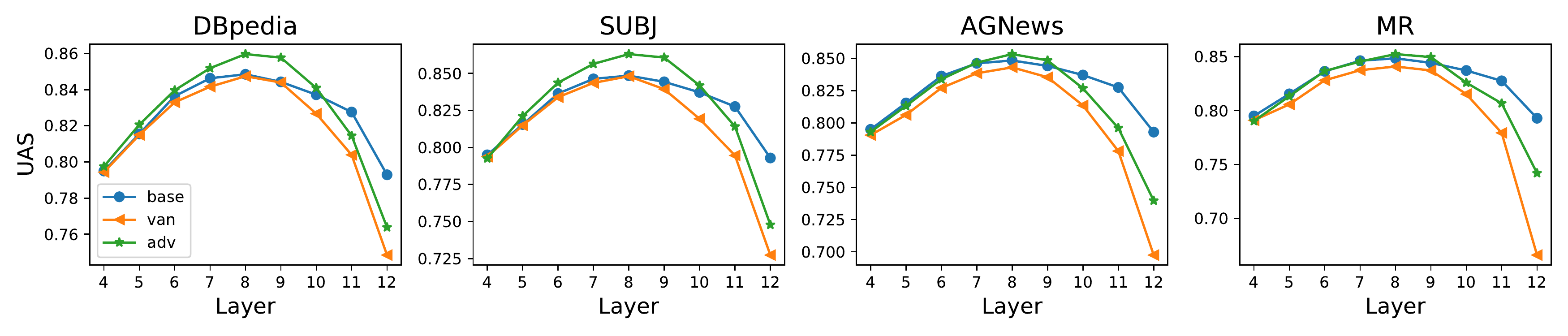}
    }
    \caption{The performance of linear probes using representations at each layer. The performance of the models degrades at higher layers, with the ADV models performing better than their VAN counterparts. The base model's plot is repeated on the four figures.}
    \label{fig:parse}
\end{figure*}
\begin{figure*}[h!]
    \centering
    \scalebox{1.0}{
    \includegraphics[width=\textwidth]{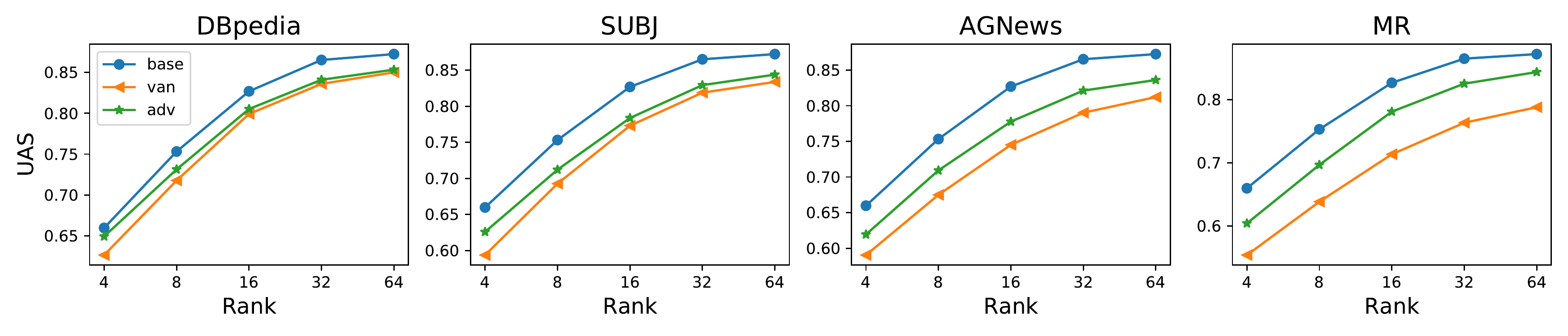}
    }
    \caption{The Pareto-based comparison of the MLP-based probes for dependency parsing using the embeddings from the last layer of the models. ADV models Pareto-dominate the VAN models. The base model's plot is repeated on the four figures.}
    \label{fig:parse-pareto}
\end{figure*}

Table \ref{tab:param-free-probe} compares the performance of four ADV and VAN models with BERT base model across two tasks from \cite{marvin2018targeted} and six tasks from \cite{warstadt2020blimp}.
As can be seen, ADV models perform better than VAN models, except for the DBpedia dataset, on which the two models perform similarly. The improvement of ADV over VAN is $21\%$ in the case of anaphora agreement on the first corpus when the models are fine-tuned on the SUBJ dataset. This number goes up to $38\%$ for the model trained on the AGNews dataset. As another example, on the irregular form task in the second corpus, the improvement over the VAN model on the MR dataset is close to $12\%$. In most of the cases, the best performing model is the base model itself, and in cases wherein a fine-tuned model does better, the difference between the base model and the winning model is often small. The only exception is the results on the Island Effects, wherein the improvements over the base model are substantial.
\subsection{Parametric Probes}
\label{sec:param}
To further contrast ADV and VAN models, we use parametric probes with linear and MLP-based parameterizations. Apart from the choice of MLP vs. linear, we follow similar architectures as \cite{pimentel2020pareto} for the probes. We use Adam \cite{kingma2014adam} with default parameters and a dropout \cite{srivastava2014dropout} of 0.3 rate on the input embeddings and after the nonlinearity in the case of MLP-based probes.

We first use linear probes for part-of-speech labeling (POSL) \cite{belinkov2017neural,hewitt2019designing}, dependency arc labeling (DAL) \cite{tenney2019you,hewitt2019structural}, and dependency parsing, using the embeddings given by the last layer of BERT. Table \ref{tab:parameter-based-probes} shows the results on the three aforementioned tasks using the English EWT corpus of Universal Dependencies \cite{nivre2017universal}. In all of the pairwise comparisons, the performance of the ADV model is superior to that of that of the VAN model.  

Figure \ref{fig:parse} shows the unlabeled attachment score (UAS) of linear probes for dependency parsing, trained on representations at each layer. Aligned with the findings of prior works \cite{tenney2019bert,jawahar2019does,hewitt2019designing,liu2019linguistic}, we observe layer-wise differentiation in BERT's performance, where middle layer probes perform the best for parsing. Additionally, we observe consistent advantage of ADV over VAN models. The trend is that in all cases the best UAS score is achieved at the eighth layer, wherein the ADV DBpedia model achieves 86.30 UAS score, surpassing the base model by 1.4 percentage points. After this layer, the performance degrades with the sharpest drops at the last two layers. ADV models have at least more than $1.0$ percentage points higher UAS than VAN models at the eight layer. This increases to 4.2 and 7.6 percentage points at that last layer for AGNews and MR, respectively. 

Drops in UAS at the last layers, which are sharper on the fine-tuned models, convey a rather drastic behavioral change in the models. We can see that the specialization of higher layers of a fine-tuned BERT to a specific task \cite{houlsby2019parameter} is achieved at the cost of a larger loss of syntactic abilities. However, our comparative analysis of VAN and ADV models indicates that AT helps prevent overspecialization of BERT to a  downstream task, and instead keeps some its more task-agnostic properties.

We conclude this section by performing a Pareto-based analysis of the performance of probes for dependency parsing, using last layer embeddings. Pareto-based analysis was proposed by \citet{pimentel2020pareto} to measure the trade-off between a probe's accuracy and its complexity. Furthermore, since complex and simple probes perform similarly on simple tasks, such as POSL and DAL, \citet{pimentel2020pareto} recommend using a more difficult task, such dependency parsing. We evaluate MLP-based probes for dependency parsing. Our MLP architecture is based on \cite{cao-etal-2021-low} where the complexity of the probe can be measured by the rank ($r$) of the weight matrix of the MLP layer,  i.e., \texttt{ReLU}($UV^Th$), where $U,V \in \mathbb{R}^{d \times r}$, where $d$ is the dimension of the input embeddings $h$. 

Figure \ref{fig:parse-pareto} plots the UAS score of each model, against the rank of the MLP layer, wherein we see the complete Pareto-domination of ADV models over their VAN counterparts. That is, the probes trained on the ADV model perform better than the ones trained on VAN models, regardless of the complexity of the probe. Similarly, the base model Pareto-dominates ADV models. 
\label{sec:intrinsic}
\begin{figure*}[h]
    \centering
    \scalebox{1.0}{
    \includegraphics[width=\textwidth]{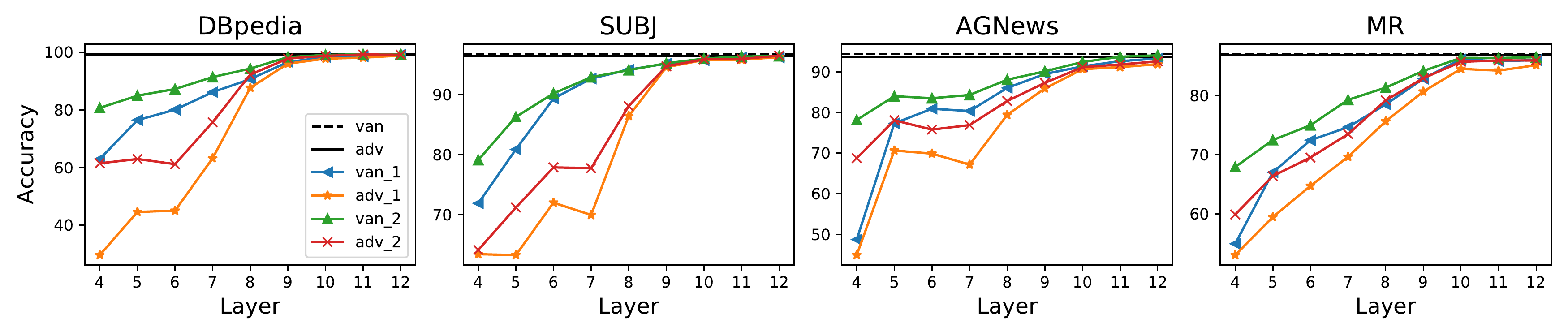}
    }
    \caption{The dev-set accuracy of models where one single layer is replaced with its rank-1 or rank-2 SVD approximation. The straight lines denote the accuracy when no approximation is done. ADV models are more sensitive to approximations and have lower accuracy throughout the network. This suggests ADV models' representations are more complex and encode more information than the VAN counterparts.}
    \label{fig:svd}
\end{figure*}
\begin{figure*}[h]
    \centering
    \scalebox{1.0}{
    \includegraphics[width=\textwidth]{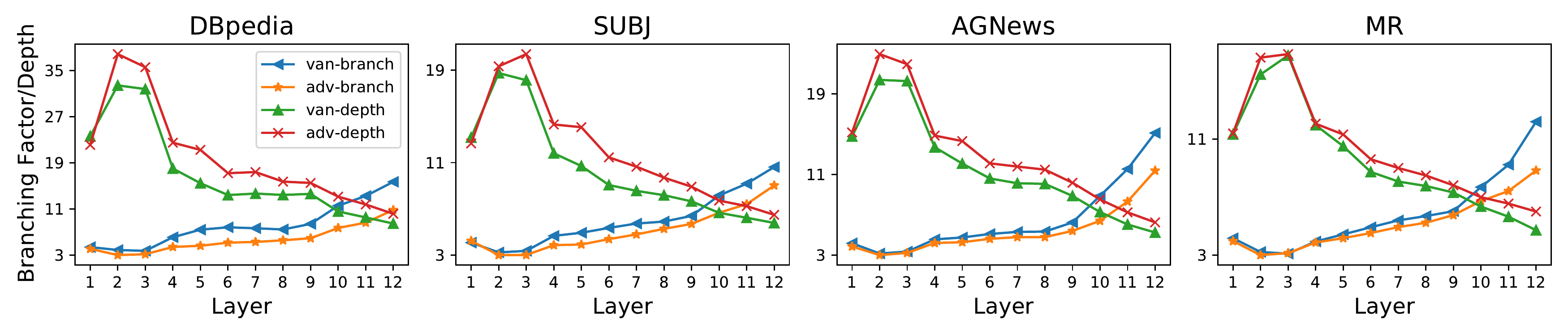}
    }
    \caption{Branching factor and depth of dependency trees given by the Chu–Liu/Edmonds' algorithm \citep{chu1965shortest,edmonds1968optimum}, averaged across all examples in the dev sets. ADV models yield smaller branching factors and larger depths.}
    \label{fig:mst}
\end{figure*}
\section{Methods for Intrinsic Analysis}
Better performances of ADV models on syntactic tasks confirm our earlier hypothesis that faithfulness to BERT's language modeling behavior and higher sensitivity to word order are good estimators of the ability of a fine-tuned model to maintain BERT's syntactic capabilities. In this section, we address \textit{how} fine-tuning affects BERT's syntactic knowledge, by introducing intrinsic measures and procedures to study sentence representations. 

We hypothesize that when a few words in a sentence/document suffice for classifying examples in a dataset, BERT tends to exhibit a bag-of-words like behavior, ignoring much of the words' structural relationships. Our hypothesis is driven by an intuition about the Transformer networks architecture \cite{vaswani2017attention}, namely the fact that the self-attention mechanism learns a convex combination of representations of all words in the sentence at every layer. This allows for some words to become more influential and dominate the sentence representation. We find that adversarial fine-tuning moderates the effect of individual words on other words, allowing for a more diverse set of word representations in a sentence.

\begin{figure*}
    \centering
    
\scalebox{0.52}{
\begin{dependency}[theme = simple]
\begin{deptext}[column sep=0.25em]
earnest \& and \& \textbf{tentative} \& even \& when \& it \& aims \& to \& shock \& . \& \\
\end{deptext}
\depedge{3}{1}{}
\depedge{3}{2}{}
\depedge{3}{4}{}
\depedge{3}{6}{}
\depedge{3}{7}{}
\depedge{3}{10}{}
\depedge{7}{5}{}
\depedge{7}{9}{}
\depedge{7}{8}{}
\end{dependency}
\begin{dependency}[theme = simple]
\begin{deptext}[column sep=0.25em]
a \& visual \& \textbf{spectacle} \& full \& of \&  {stunning} \& images \& and \& effects \& . \& \\
\end{deptext}
\depedge{3}{1}{}
\depedge{3}{2}{}
\depedge{3}{10}{}
\depedge{6}{4}{}
\depedge{9}{6}{}
\depedge{3}{5}{}
\depedge{6}{7}{}
\depedge{9}{8}{}
\depedge{3}{9}{}
\end{dependency}
\begin{dependency}[theme = simple]
\begin{deptext}[column sep=0.25em]
the \& soundtrack \& alone \& is \& worth \& the \& \textbf{price} \& of \& admission \& . \& \\
\end{deptext}
\depedge{2}{1}{}
\depedge{7}{10}{}
\depedge{9}{8}{}
\depedge{7}{9}{}
\depedge{7}{2}{}
\depedge{7}{3}{}
\depedge{7}{4}{}
\depedge{7}{5}{}
\depedge{7}{6}{}
\end{dependency}
}
    \caption{Three examples of dependency trees extracted from gradients of the ADV model at the last layer's representations. The root in each example is denoted by the bold font. The depth of the trees are 2,3, and 2, from left to right, respectively. In contrast, the VAN model produce depth-1 trees (not shown in the figure) with the roots (tentative, stunning, and price) directly connected to all the other nodes.}.
    \label{fig:deptrees}
\end{figure*}

\begin{figure*}[h!]
    \centering
    \scalebox{1.0}{
    \includegraphics[width=\textwidth]{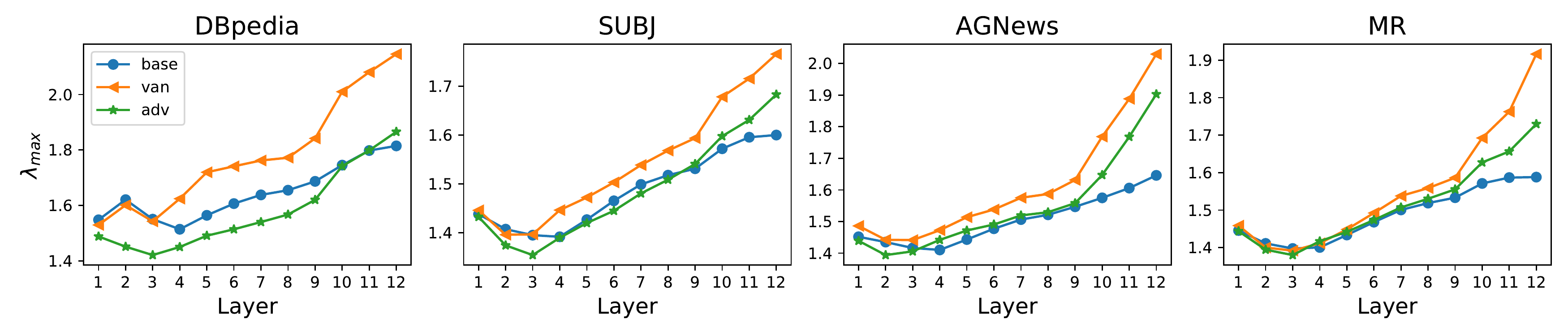}
    }
    \caption{Largest eigenvalue ($\lambda_{max}$) of the Laplacian of the symmetrized influence graph at each layer. ADV models are more similar to the base model and have smaller $\lambda_{max}$ compared with VAN models. The plots for all models are created based on the examples in the dev sets.}
    \label{fig:connectivity}
\end{figure*}
 \subsection{SVD-Based Analysis}
As one or a few words become more dominant and affect other words, we expect the matrix representing the sentence will lean towards a more low-rank matrix. We inspect the performance of the models, given low-rank approximations of hidden layers, as a measure of the richness or complexity of the representations. Concretely, we use a rank-$r$ approximation of the representation matrix and replace the $l_\text{th}$ hidden layer, $h^l$, with 
\begin{equation*}
    \text{SVD}_r(h^l) = \sum_{i=1}^rU_i \sigma_i  V_i^t
\end{equation*}
wherein $U_i, \sigma_i,$ and $V_i$ are the $i_\text{th}$ left singular vector, singular value, and the right singular vector, associated with the singular value decomposition of $h^l$. We pass this low-rank approximation to the next layer, keep everything else intact, and measure the accuracy on the dev set for each dataset. 

In Figure \ref{fig:svd}, we plot the accuracy at the $i_{\text{th}}$ layer by replacing  BERT's output with the following,
\begin{equation*}
    L_{n}\Bigg(... L_{i+1}\bigg(\text{SVD$_{r}$}\Big(L_{i}\big(... L_{1}(x)\big)\Big)\bigg)\Bigg)
\end{equation*}
where $L_i$ denotes the $i_{\text{th}}$ layer of BERT. While the two models have very similar accuracies using the full-rank matrices, the ADV models have lower accuracies using the approximations. This is more evident in middle layers, where the two models also start to demonstrate divergent syntactic performance, shown earlier in Figure \ref{fig:parse}. Severe degradation in the performance suggests ADV models' representations are more complex and encode more information than their VAN counterparts.

Theoretical machine learning works have shown a \textit{simplicity} bias in learned functions \cite{valle2018deep} and learned weights \cite{huh2021low} of neural networks. Our SVD-based measure shows that BERT tends to simplify/compress representations towards the higher layers. AT mitigates oversimplification by maintaining more complex representations. 

\subsection{Graph-Based Analysis}
In order to measure how certain words dominate other words' representations, we use the following gradient-based measure, which gives us a pairwise graph of how words influence each other. 
\begin{equation}
\forall i \neq j  \quad S^{l}_{ij} = \lVert \frac{\partial \lVert h^l_{i}\rVert}{\partial x_j} \rVert
\end{equation}
The score $S^{l}_{ij}$ estimates the influence of the j$_{\text{th}}$ word on the representation of the i$_{\text{th}}$ word at the l$_{\text{th}}$ layer. 
This metric has also been used in a concurrent work \cite{meister2021sparse} which studies the instance-level interpretability of models' predictions. We, instead, study the properties of the graphs induced by this metric.
 
For our first graph-based analysis, we pick the $j_\text{th}$ word with the highest sum of influence at each layer, i.e., $\text{argmax}_{j}(\sum_{i}S^{l}_{ij})$, as the root. We then normalize the scores such that $\forall i, \sum_{j}S^{l}_{ij}=1$\footnote{We found normalization to have virtually no effect on the results. Nevertheless, since our next graph-based analysis requires normalization, we used normalized scores for this analysis too.}. Finally, we find a \textit{spanning arborescence} of maximum weight which is the directed graph analogue of the maximum spanning tree problem (MST). We use the Chu–Liu/Edmonds' algorithm \citep{chu1965shortest,edmonds1968optimum} which has also been used for non-projective dependency parsing \cite{mcdonald2005non}. 
Figure \ref{fig:mst} contrasts the branching factor (maximum outdegree) and the depth of the extracted dependecy trees of ADV and VAN models, averaged across all examples in the dev sets. As can be seen, ADV models maintain a smaller branching degree and a larger depth across all layers of the network. Branching factor of a tree indicates the reach of its most influential nodes. Conversely, a more moderate branching factor leads to deeper trees.

This graph-based analysis gives us a clearer picture of the difference between the two models. Specifically, larger branching factors in VAN models suggest that they tend to overestimate the importance of some words (e.g., sentiment-related words in sentiment classification), leading to less representational diversity. Consequently, their smaller depths indicate weaker sensitivity to hierarchies, leading to the degradation of syntactic abilities. Figure \ref{fig:deptrees} depicts three extracted dependency trees using the ADV model trained on the MR dataset. The depth of the tree in all of these cases is larger than 1. In contrast, VAN models produce flat structures with depth 1, in which the root is the only head in the sentence and is directly connected to all the other nodes in the tree. 
 
As a complementary perspective, we study a spectral property of the induced graphs to highlight structural differences among VAN and ADV models. Formally, given an undirected graph $G(V,E)$, the graph Laplacian is given by
\begin{equation*}
    L = D - A
\end{equation*}
where $A$ is the adjacency matrix\footnote{We symmetrize the influence graph by means of adding it to its transpose.} and $D$ is a diagonal matrix, wherein each diagonal elements is the sum of the outgoing edge weights from a node, i.e., $D_{ii} = \sum_{j}A_{ij}$. The Laplacian is a positive semi-definite matrix and its spectrum is elegantly associated with connectivity properties in graphs \cite{fiedler1973algebraic,oellermann1991laplacian}.

Here, we use the largest eigenvalue of the Laplacian due to its connection to the problem of finding the maximum-cut in a graph \cite{mohar1990eigenvalues}. The objective in the maximum-cut problem is to find a cut $S \subset V$ that maximizes the total weight of the edges between $S$ and $V-S$. This is known to be an NP-hard problem \cite{garey1979computers} for which only approximate algorithms exist. As such, \citet{mohar1990eigenvalues} give $\frac{|V|}{4}\lambda_{max}$ as the upper bound on the value of the maximum cut, where $\lambda_{max}$ denotes the largest eigenvalue of the Laplacian. We use the max cut property due to its relevance to our finding on the geometry of the induced dependency trees, where we observed increasing branching factors across layers. Specifically, as a few nodes dominate the influence graph, we expect an approximate max cut to be achieved by selecting the cut separating influential nodes from others. 

Figure \ref{fig:connectivity} depicts the average of the largest eigenvalue of influence graphs, across all examples in dev sets, at each layer, wherein we see larger values in higher layers. Since influence graphs are already normalized, the increase in $\lambda_{max}$ highlights a change in the geometry of the graphs, rather than a simple change in the magnitude of the weights of the graphs. Further, we plot the values for the based model and observe a larger divergence for VAN models on all datasets. Indeed, on all datasets except DBpedia, we see increased divergences for the fine-tuned models at the last three layers, similar to our observations on parametric probes in Figure \ref{fig:parse}. This spectral analysis, too, reveals inherent differences among induced representations of ADV and VAN models, in a manner well-aligned with our previous analyses.
\subsection{Discussion}
We have proposed new methods of analysis which aim to complement the probe-based analysis of language models covered in \S \ref{sec:free} and \S \ref{sec:param}. Crucially, our intrinsic methods can provide insights about the inner workings of BERT independently of linguistic formalisms, albeit only for the fine-tuning phase. We take a step towards a better understanding at how BERT's higher layers become more task-specific, leading to its loss of syntactic abilities, and how adversarial training mitigates this. The key factor enabling such analysis is our choice of contrastive analysis of two training approaches which bring about clearly different structural properties in representations. By contrasting vanilla training with adversarial training, we not only explain the benefits of adversarial training, we also add methods to the NLP interpretability toolbox which could be useful for the general analysis of neural models. 
\section{Related Work}
\label{sec:related}
Adversarial training has been found to be effective for regularizing language models (LMs) during fine-tuning \cite{jiang2019smart,zhu2019freelb,pereira2021targeted}, as well as pretraining \cite{liu2020adversarial}. Few recent works have studied the dynamics of adversarial training theoretically using two-layer MLP models \cite{awasthi2020adversarial,allen2020feature}. We propose a comparative analysis of vanilla training and adversarial training to find a linguistic interpretation for adversarial training. 
 
Interpreting neural network models has been the subject of numerous research papers in the past few years. One class of interpretability research focuses on the direct examination of LMs, regarding what and how they learn \cite{rogers2020primer}. The consensus has been that BERT, which is widely studied, has a remarkable syntactic knowledge, with the middle layers performing the best on syntax-related tasks \cite{liu2019linguistic,tenney2019bert,jawahar2019does,hewitt2019designing}. Nevertheless, the way this knowledge is affected after fine-tuning is not explored.   
Recent papers \cite{mosbach2020stability,zhang2020revisiting,aghajanyan2020intrinsic} study the dynamics of fine-tuning from an optimization perspective. However, to the best of our knowledge, no prior work has studied syntactic abilities of fine-tuned models.  
 
Probe classifiers require extra parameters, which could be confounding factors in the analysis of the linguistic capabilities of a models; i.e., to what extent the performance of the probe relies on the model's representations. This has been the subject of several recent works which aim to measure the true capabilities of LMs \cite{voita2020information,pimentel2020pareto,hewitt2019designing}. Recent works interpret LMs without the use of a probe classifier \cite{wu2020perturbed,chrupala2019correlating,cao-etal-2021-low}. Most notably, \citet{wu2020perturbed} extract dependency trees using a two-phase procedure which requires masking pairs of words from text. We use a gradient-based measure for the same purpose, but one which can be computed more directly without the need for masking. Furthermore, we use our gradient-based procedure for an intrinsic analysis of BERT's syntactic representation, as opposed to mapping it to linguistic formalisms. 
 \section{Conclusion and Future Work}
 \label{sec:conclusion}
 We uncover a major reason behind empirical benefits \cite{zhu2019freelb,pereira2020adversarial,pereira2021targeted,jiang2019smart,aghajanyan2020better} of adversarial fine-tuning of BERT. We demonstrate that adversarial fine-tuning moderates BERT's loss of syntactic abilities and helps with creating more task-agnostic representations. We use standard probe-based techniques and provide new methods of analysis to explain the benefits of adversarial training. We find vanilla-trained models to oversimplify sentence representations, causing the model to be more resistant to input corruption by low-rank approximations. We also study geometric properties of the representations and find that vanilla-trained models tend to overestimate the importance of a few words which dominate sentence representations, drifting the fine-tuned model further from the base BERT model.    
 
 We expect that recent regularization schemes \cite{aghajanyan2020better,cheng2020posterior} and multi-task learning \cite{liu2019multi} provide benefits similar to the ones we presented for adversarial training in this paper. However, none of the existing methods directly optimize for syntactic generalization. Syntactic abilities of LMs are at the core of their success, and thus, future work should create better regularization schemes which directly preserve syntactic abilities. Nonetheless, maintaining rich structured representations could be at odds with other goals of a user, such as feature-attribution interpretability where sparsity is desired \cite{ribeiro2016should,de2020decisions}. Therefore, characterizing neural models' representations from a multitude of aspects and the studying of their interplay are important directions for future research.

\bibliography{tacl2018}
\bibliographystyle{acl_natbib}

\end{document}